\pdfoutput=1
\documentclass[11pt]{article}

\usepackage[preprint]{acl}

\usepackage{times}
\usepackage{latexsym}
\usepackage{float}
\usepackage[T1]{fontenc}

\usepackage[utf8]{inputenc}

\usepackage{microtype}

\usepackage{inconsolata}

\usepackage{graphicx}

\usepackage{amsmath,amsfonts,bm}

\def\eqref#1{equation~\ref{#1}}

\def\1{\bm{1}}

\DeclareMathAlphabet{\mathsfit}{\encodingdefault}{\sfdefault}{m}{sl}
\SetMathAlphabet{\mathsfit}{bold}{\encodingdefault}{\sfdefault}{bx}{n}

\def\gD{{\mathcal{D}}}

\def\gQ{{\mathcal{Q}}}

\usepackage{inconsolata}

\usepackage{arydshln}

\usepackage{graphicx}
\usepackage{multirow}
\usepackage{multicol}
\usepackage{amssymb}
\usepackage{amsmath}
\usepackage{tabulary}
\usepackage{booktabs}
\usepackage{enumitem}
\usepackage{subcaption}
\usepackage{wrapfig}
\usepackage{xcolor}
\usepackage{longtable}
\usepackage{makecell}
\usepackage{pdfpages}

\usepackage[utf8]{inputenc}
\usepackage{lipsum}  
\usepackage{enumitem}
\usepackage{tcolorbox}
\usepackage{amssymb} 
\usepackage{fontawesome} 

\usepackage{listings}
\usepackage{xcolor} 
\usepackage{amsthm}
\usepackage{tcolorbox}

\usepackage{soul,color,xcolor}      
\definecolor{myColor}{RGB}{0,0,200}   
\makeatletter
\newcommand*{\new}{\@ifnextchar\bgroup{\new@}{\color{myColor}}}
\newcommand*{\new@}[1]{{\textcolor{myColor}{#1}}}
\makeatother

\usepackage{xspace}
\newcommand{\method}{\textsc{Active-Critic}\xspace}

\usepackage{hyperref}
\newcommand{\Sref}[1]{\S\ref{#1}}
\usepackage{array}

\makeatletter
\def\adl@drawiv#1#2#3{%
        \hskip.5\tabcolsep
        \xleaders#3{#2.5\@tempdimb #1{1}#2.5\@tempdimb}%
                #2\z@ plus1fil minus1fil\relax
        \hskip.5\tabcolsep}
\newcommand{\cdashlinelr}[1]{%
  \noalign{\vskip\aboverulesep
           \global\let\@dashdrawstore\adl@draw
           \global\let\adl@draw\adl@drawiv}
  \cdashline{#1}
  \noalign{\global\let\adl@draw\@dashdrawstore
           \vskip\belowrulesep}}
\makeatother

\title{Large Language Models Are Active Critics in NLG Evaluation}

\author{
    Shuying Xu\\
    \small Tongji University\\
    \small \texttt{shuyingxu@tongji.edu.cn}
    \And
    Junjie Hu \\
    \small University of Wisconsin-Madison \\
    \small \texttt{junjie.hu@wisc.edu}\\
    \And
    Ming Jiang\thanks{Corresponding author.}\\
    \small Indiana University Indianapolis \\
    \small \texttt{mj200@iu.edu}
}

\begin{document}
\maketitle

\begin{abstract}
The conventional paradigm of using large language models (LLMs) for natural language generation (NLG) evaluation relies on pre-defined task definitions and evaluation criteria, positioning LLMs as ``passive critics'' that strictly follow developer-provided guidelines. However, human evaluators often apply implicit criteria, and their expectations in practice can vary widely based on specific end-user needs. Consequently, these rigid evaluation methods struggle to adapt to diverse scenarios without extensive prompt customization. To address this, we introduce \textbf{\method}, a novel LLM-based evaluator that transforms LLMs into ``active critics'' capable of adapting to diverse NLG tasks using limited example data. \method consists of two stages: (1) self-inferring the target NLG task and relevant evaluation criteria, and (2) dynamically optimizing prompts to produce human-aligned scores along with detailed justifications. Our experiments show that \method can generate nuanced, context-aware evaluation criteria, enabling it to achieve superior alignment with human judgments across multiple tasks.

\end{abstract}

\section{Introduction}
\label{sec:introduction}

Recent advances in language technologies have accelerated the development of natural language generation (NLG) systems, benefiting a variety of downstream applications such as text summarization~\citep{fabbri2021summeval}, dialogue generation~\citep{mehri2020topicalchat}, and storytelling~\citep{guan2021openmeva}. However, despite the rapid progress in NLG systems, reliable techniques for automatic evaluation of NLG systems still lay far behind, primarily due to the inherent challenges posed by the open-ended nature of NLG and the diverse demands of different stakeholders. This gap, in return, undermines the reliability of machine-generated content in real-world applications.

Traditional NLG evaluation methods typically focus on a specific criterion and require human-written references for comparison~\citep{li2024leveraging}. Commonly considered criteria include reference similarity~\citep{papineni2002bleu,lin2004rouge,zhang2019bertscore,yuan2021bartscore}, text fluency~\citep{fluency1,fluency2}, human likeness~\citep{humanlikeness1,humanlikeness2}, and information adequacy~\citep{correctness1}. Moving beyond single-aspect metrics, recent studies propose to use a universal large language model (LLM) as a judge to score machine-generated texts across multiple criteria in diverse NLG tasks, either by fine-tuning ~\citep{zhong2022unieval,jiang2023tigerscore,xu2023instructscore,ke2023critiquellm} or by prompting an LLM for assessment~\citep{chiang2023can,gong2023coascore,lin2023unlocking}. To address the high cost of human annotation and potential biases introduced by limited references, researchers have further developed reference-free LLM-based evaluations~\citep{fu2023gptscore,G-eval2023g,li2023autoj,jia2023zero}. 

Despite the remarkable advancements of prior work, one major concern remains: the reliance on \textit{pre-defined evaluation task descriptions and criteria} for assessment, forcing LLM evaluators to adhere strictly to developers' expectations. In contrast, human evaluators often use nuanced, implicit evaluation criteria that extend beyond these pre-defined criteria in practice~\citep{liu2024hd,clark2021human-not-gold,celikyilmaz2020evaluation-survey}. While recent studies~\citep{liu2024hd,li2024decompose,liu2023calibrating} have explored prompting LLMs to generate evaluation criteria automatically, these methods still rely on pre-defined task descriptions, requiring substantial manual effort to tailor prompts for each NLG task. Moreover, even within the same task, different stakeholders may prioritize distinct evaluation criteria, making it potentially risky to establish a fixed set of criteria in advance without first accounting for human evaluation nuances. 

To overcome the above limitations, we propose a novel evaluation approach, i.e., \textbf{\method}, that instructs an LLM to actively derive an evaluation protocol purely from human-scored data examples. Our approach includes two stages: (1) adaptively inferring the target NLG task and identifying its underlying evaluation criteria that matter most to end users, and (2) dynamically optimizing prompts to produce human-aligned judgments across diverse NLG scenarios. To enhance trustworthiness, \method also generates detailed text justifications alongside its scoring. 

We have conducted experiments across diverse NLG tasks using various base LLMs. The results show that the \method consistently achieves a noticeably higher correlation with human judgments, indicating its ability to adapt effectively to different NLG evaluation tasks according to different evaluation criteria. Our approach requires as few as 5 human-scored data to obtain a strong correlation with humans, with performance steadily improving as the dataset grows. Further analysis highlights that the task inference stage contributes more to \method's performance than the scoring stage, and \method can effectively identify nuanced, context-aware criteria beyond pre-defined ones. In summary, our method offers three key benefits: 

\begin{itemize}[leftmargin=13pt]
    \item \textbf{Self-adaptive evaluation.} \method can infer any NLG evaluation task, recover human judgment criteria, and make justified assessments directly from data, eliminating the need for pre-defined task descriptions, fixed evaluation criteria, or manual prompt engineering. 
    \item \textbf{Accurate judgment alignment.} Our two-stage design guides LLMs to mimic human judgment step by step, yielding interpretable justifications while achieving state-of-the-art alignment with human assessments against strong baselines. 
    \item \textbf{Generic for diverse LLMs and NLG tasks. } Our method operates independently of specific LLMs and evaluation tasks. Our results on four LLM backbones across four NLG tasks showcase its broad applicability.
\end{itemize}

\section{Related Work}
\label{sec:relatedwork}

\paragraph{NLG Evaluation Overview.} Existing methods for NLG evaluation span three major strands, including early human-centric evaluation~\citep{mellish1998huamnevaluation}, followed by untrained machine evaluation~\citep{papineni2002bleu,lin2004rouge,lavie2009meteor}, and more recently, machine-learned evaluation~\citep{sennrich2015neural,zhang2019bertscore,yuan2021bartscore,kim2023prometheus}. These studies largely concentrate on single-criteria metric design, targeting either general NLG tasks like reference alignment~\citep{liu2023Reference-Based} or a specific NLG task like coherence for text summarization~\citep{wang2023element}. To enhance evaluation efficiency, recent works have advocated for unified evaluation frameworks built upon LLMs, aiming to transcend task-specific boundaries and assess multiple criteria simultaneously~\citep{chiang2023can,liu2023calibrating,gong2023coascore,liu2024hd,li2024decompose}. Our work falls into this group, and will discuss the details below.

\paragraph{LLM-based NLG Evaluation. } Prior studies on unified evaluation frameworks primarily focus on enhancing evaluation generalizability, with an emphasis on estimating instance quality scores across various NLG tasks and multiple criteria simutaneously~\citep{xiao2023MetricEva,gao2024challenges}. Studies in this area typically involve two strategies. One is developing criteria-centered prompts that guide LLMs as a judge for multi-faceted, train-free evaluations~\citep{fu2023gptscore, G-eval2023g,lin2023llmeval,chiang2023closerlook,li2024decompose,liu2024hd,yuan2023batcheval}. The other focuses on curating a large-scale multi-scenario benchmark to fine-tune an LLM as a generalized evaluator~\citep{zhong2022unieval,li2023autoj,wang2023pandalm,ke2023critiquellm,kim2023prometheus,hu2024themis}. 

While prompting-based methods are more cost-effective than tuning-based ones, one major concern with these approaches is the sensitivity of LLMs to manual prompts, possibly causing evaluation biases. To address this issue, several latest works have explored instructing LLMs to generate evaluation criteria ~\citep{liu2024hd,li2024decompose} or scoring rubrics~\citep{liu2023calibrating} based on pre-defined context like the target NLG task description. In essence, criteria generation in these studies implicitly assumes that each NLG task has a fixed set of evaluation criteria. In contrast, we argue that different end-user needs may lead to varying emphases, even for the same NLG task, resulting in criterion and/or rubric variation. To address this, our approach takes a data-driven perspective, instructing the LLM for NLG evaluations through self-inference of all relevant contexts.

\paragraph{Dynamic Prompt Optimization. } 
Dynamic prompt optimization iteratively refines prompts to enhance the performance of static LLMs on specific tasks. Existing methods can be divided into two categories based on their inference depth. Single-layer optimization methods, such as APE~\citep{zhouAPE}, APO~\citep{pryzant2023APO}, OPRO~\citep{yang2023OPRO}, and IPC~\citep{levi2024IPC}, focus on optimizing prompts within a single stage, limiting their adaptability to complex tasks.  In contrast, multi-layer optimization methods, like DSPy~\citep{khattab2023dspy} and MIPRO~\citep{opsahl2024MIPRO}, refine prompts across multiple stages, supporting more comprehensive reasoning but relying on scalar-based comparisons between data points, which are insufficient for tasks requiring correlations across data vectors. We design a correlation-based comparison rather than a scalar-based one to optimize multi-stage NLG evaluation tasks. 
\section{Notations and Problem Definition}
\label{sec:prelim}
Our goal is to develop a highly adaptive NLG evaluation approach that can dynamically align with diverse end-user preferences to make explainable judgments across diverse NLG scenarios. Specifically, given a small set of source-response-quality tuples $\mathcal{D} = \{(x_i, y_i, r_i)\}_{i=1}^N$ annotated by humans based on their hidden criteria $\mathcal{C}=\{c_1, ..., c_k\}$, we aim to build an LLM-based reference-free evaluator $E(x', y')$. This evaluator learns from the annotated dataset $\mathcal{D}$ to infer task-relevant information, including the target NLG task description $T$ and the evaluation criteria $\hat{\mathcal{C}} = \{\hat{c}_1, \dots, \hat{c}_m\}$. Using this inferred information, it can estimate the quality score $\hat{r}$ of the source-response pair $(x', y')$, along with a free-text justification $\hat{e}$. Here, $x_i$ denotes the $i$-th input text from the original NLG task, while $y_i$ denotes the corresponding response generated by an NLG system and $r_i$ is the quality score of $y_i$. We denote $\text{LLM}(\texttt{[prompt]})\to \texttt{[response]}$ as the response generation by LLM given a prompt.

\section{\method}
\label{sec:methods}

\begin{figure*}[t]
    \vspace{-5mm}
    \centering
    \includegraphics[width=\linewidth]{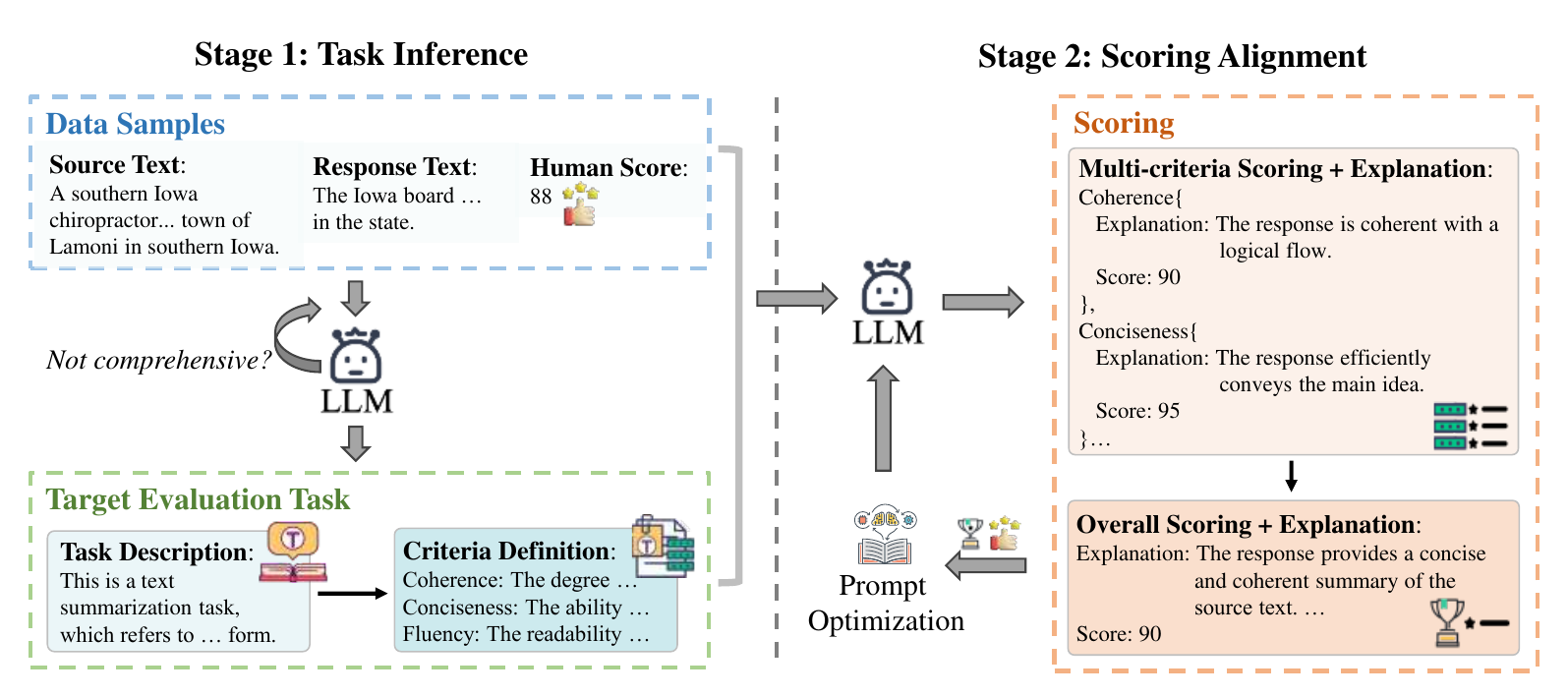}
    \caption{Overview of \method, including two stages: (1) task inference, where the LLM is instructed to derive the target NLG evaluation task description and relevant criteria from data samples, and (2) scoring alignment, allowing the LLM to generate multi-criteria and overall quality scores along with accompanying explanations.}
    \label{fig:workflow}
\end{figure*}

\paragraph{Overview.} Figure~\ref{fig:workflow} shows the overall workflow of \method. With the motivation that an ideal unified evaluation framework should flexibly uncover the nuanced evaluation criteria of end users across diverse generation scenarios and make human-aligned judgments, we design a data-driven evaluation framework structured in two stages. The first stage is \textit{task inference} (\Sref{sec:method:taskinference}), where we instruct an LLM to predict task-related information by actively reviewing a small set of human-rated data examples. Through this analysis of the human-rated data, we expect the model to self-infer the details of the target evaluation task and the implicit criteria used by human annotators. The second stage is \textit{scoring alignment} (\Sref{sec:method:protocolopt}), where we aim to align the LLM evaluator with human scoring based on the predicted evaluation criteria. Specifically, we design a dynamic prompt optimization method to automatically select the optimal few-shot examples, $\gD_\text{demo}$, from $\mathcal{D}$ which enables the LLM evaluator to achieve human-aligned scoring through in-context prediction. 

\subsection{Task Inference}
\label{sec:method:taskinference}
The task inference stage, depicted on the left side of Figure~\ref{fig:workflow}, focuses on identifying two key components for NLG evaluation: (1) \textit{task description} and (2) \textit{criteria definition}. This stage uses the LLM to analyze the dataset $\mathcal{D}_\text{train}$, infer the characteristics of the NLG task, and establish relevant evaluation criteria without human intervention. 


\paragraph{Task Description.} This module instructs the LLM to formulate an accurate task description $T$ by reviewing examples in $\mathcal{D}_\text{train}$ and identifying key information that characterizes the target NLG task (e.g., summarization, storytelling) for evaluation. Considering that LLM's context length limit may not fit in all examples in $\gD_\text{train}$, we split these examples into $N$ mini-batches, and generate one task description $T_n$ from each mini-batch $\gD_{\text{train},n}$. That is, 
$\text{LLM}(f_t(\gD_{\text{train},n})) \to T_n,~\forall n\in [1, N]$, 
where $f_t$ is a prompt template shown in Table~\ref{tab:appendix-TaskDescription} in the appendix. The final task description $T$ is generated by the LLM through the ensemble of all task descriptions $\{T_n\}_{n=1}^N$ over all mini-batches.

\paragraph{Criteria Definition.} After establishing the task description for each mini-batch, the LLM is instructed to define task-specific evaluation criteria for assessing the quality of machine-generated texts. Unlike traditional evaluation frameworks that rely on predefined criteria (e.g., coherence, fluency), we instruct the LLM to automatically identify the most relevant evaluation dimensions for the target NLG task. Similar to the task description, the final criteria set $\hat{C}=\{\hat{c}_1, \hat{c}_2,\dots,\hat{c}_m\}$\footnote{We instruct the LLM to output a criteria set in the JSON format, as shown in Table~\ref{tab:appendix-CriteriaDefinition} in the appendix.} is composed of all relevant dimensions inferred from all the reviewed mini-batches.

To enhance efficiency, we instruct the LLM to decide whether to stop early based on the comprehensiveness of the generated task description and criteria set after processing each mini-batch.

\subsection{Scoring Alignment}
\label{sec:method:protocolopt}

Our second stage, as shown on the right side of Figure~\ref{fig:workflow}, focuses on aligning the LLM evaluator with human scoring judgments by automatically optimizing the evaluation prompts. Inspired by prior research that harnesses the potential of LLMs by breaking down complex tasks into simpler ones~\citep{wei2022chain-of-thought,khotdecomposedPrompting}, we hypothesize that starting with fine-grained, criteria-specific scoring can help the model further derive an accurate overall quality score. With this intuition in mind, we structure the scoring stage into two modules: (1) \textit{Multi-criteria Scoring with Explanation (McS-E)}, followed by (2) \textit{Overall Scoring with Explanation (OS-E)}. 

\paragraph{Multi-criteria Scoring with Explanation (McS-E). } 
In this module, we use the LLM to assess the model output $y_i$ based on the criteria set $ \hat{C} = \{\hat{c}_1, \hat{c}_2, \dots, \hat{c}_m\} $ obtained from the \textit{task inference} stage (\Sref{sec:method:taskinference}). Specifically, for each input-output pair $(x_i, y_i)$, the LLM is instructed to estimate a score $\hat{r}_{ij}$ and a corresponding explanation $e_{ij}$ according to each criterion $\hat{c}_j\in \hat{C}$:
\begin{align}
    &\text{LLM}(x_i,y_i,f_\text{McS-E}(T,C,\gD_\text{demo})) \to \hat{R}_i \\
    &\hat{R}_i = \{ (\hat{r}_{ij}, \hat{e}_{ij}),~\forall \hat{c}_j \in \hat{C}\}
\end{align}
where the output uses a JSON format, indicating a set of score-explanation pairs $\hat{R}_i$ for all criteria in $\hat{C}$ and $\gD_\text{demo}$ is a set of demonstration examples randomly selected from the training set $\gD_\text{train}$. This mechanism ensures that the evaluation is both quantitative and interpretable, offering insights into the rationale behind each score. The prompt template $f_{\text{McS-E}}(T,C,\gD_\text{demo})$ is designed to enable scoring across multiple criteria simultaneously, accounting for the interconnections between them. This design enables a fine-grained evaluation, where each criterion is treated both individually and in connection with the others, providing detailed explanations that enhance the interpretability of the scoring process.

\paragraph{Overall Scoring with Explanation (OS-E). } 
After scoring the individual criteria, we use a prompt template $f_\text{OS-E}$ to instruct 
the LLM to synthesize these scores $\{\hat{r}_{i1}, ...\hat{r}_{im}\}$ into an overall quality score $\hat{r}_{i}$, and an explanation $e_{i}$ that provides a comprehensive justification for the final decision. 
\begin{equation} \label{eq:Overall Quality Scoring}
\text{LLM}(x_i,y_i, f_\text{OS-E}(T,\hat{R}_i,\gD_\text{demo})) \to \hat{r}_i, \hat{e}_{i}
\end{equation}


\paragraph{Prompt Optimization.} Given the sensitivity of LLMs' in-context prediction performance to the few-shot examples $\gD_\text{demo}$ in the prompt, we further propose an automatic prompt optimization strategy built upon DSPy~\citep{khattab2023dspy} to iteratively select the optimal $\gD_\text{demo}^*$ to refine the prompts. Specifically, given two lists of overall quality scores across all examples in $\gD_\text{train}$---one predicted by the LLM, i.e., $\hat{r}=[\hat{r}_1, \dots, \hat{r}_N]$ from Eq.~(\ref{eq:Overall Quality Scoring}), and the other annotated by humans, i.e., $r=[r_1, \dots, r_N]$---we design an objective function to maximize the correlation between these two score lists. To mitigate potential biases caused by relying on a single correlation measurement, we calculate the sum of three widely-used correlation coefficients: Pearson ($\gamma$), Spearman ($\rho$), and Kendall ($\tau$) with equal weights:
\begin{align}\label{eq:validation-metric}
    \mathcal{Q}(\hat{r},r) &= \gamma(\hat{r},r) + \rho(\hat{r},r) + \tau(\hat{r},r) \\
    \gD_\text{demo}^* &= \arg \max_{\gD_\text{demo} \subset \gD} \gQ(\hat{r}, r) 
\end{align}
where $\gD_\text{demo}$ is the optimal few-shot demonstration examples $\gD_\text{demo}$ selected from $\gD_\text{train}$. To approximately solve the above maximization problem, we repeat $K$ time for the evaluations of Eq.~(\ref{eq:Overall Quality Scoring}) using different randomly sampled $\gD_\text{demo}$, and select the best $\gD_\text{demo}^*$ that maximizes $\gQ(\hat{r}, r)$.  
\section{Experiment Settings}
\label{sec:settings}

\paragraph{Benchmarks}

Following prior work~\citep {zhong2022unieval,fu2023gptscore,G-eval2023g}, we evaluate our method on four popularly-used benchmarks. These datasets cover diverse topics (e.g., politics, sports, restaurants, etc.) across four NLG tasks (i.e., summarization, dialogue generation, data-to-text generation, and storytelling), aiming to construct a robust testbed to access \method. The details of each benchmark are described below.

\begin{itemize}[leftmargin=6mm]
    \item \textbf{SummEval}~\citep{fabbri2021summeval}: 1,600 machine-generated summaries of CNN/DailyMail articles were rated by both expert and layman judges on coherence, consistency, fluency, relevance, and overall quality.

    \item \textbf{Topical-Chat}~\citep{mehri2020topicalchat}: A knowledge-grounded, open-domain dialogue dataset consisting of 60 conversations, each paired with 6 responses (2 by humans and 4 by machines). Responses are human-evaluated on overall quality across five dimensions: naturalness, coherence, engagingness, groundedness, and understandability.

    \item \textbf{SFRES}~\citep{wen2015sfres}: A data-to-text generation benchmark with 1,181 instances, focusing on generating free-text utterances from structured restaurant information. Annotators rated the overall quality of each instance based on informativeness and naturalness.

    \item \textbf{OpenMEVA (ROC)}~\citep{guan2021openmeva}: 1,000 open-ended commonsense stories generated by various models trained upon the ROCStories corpus. Annotators rate each story based on fluency, creativity, and coherence.

\end{itemize}

We standardize all benchmarks into a uniform format that includes: (1) the machine-generated responses for evaluation, (2) the source input used by the generation systems for response generation, and (3) the human scores assessing response quality.

\paragraph{Baselines and Metrics}
We compare \method with a variety of state-of-the-art publicly accessible NLG evaluation methods. The baselines are grouped into two categories: (1) fine-tuning-based methods including Auto-J ~\citep{li2023autoj}, UniEval~\citep{zhong2022unieval}, InstructScore~\citep{xu2023instructscore} and TIGERScore~\citep{jiang2023tigerscore}; and (2) prompting-based methods, including GPTScore~\citep{fu2023gptscore}, G-eval~\citep{G-eval2023g} and four selected base LLMs under the zero-shot manner, implemented following~\citep{mahmoudi2023exploring}. Following prior work~\citep{fu2023gptscore, jiang2023tigerscore}, we use GPTScore-src to refer to the source-hypothesis scoring type, which is a reference-free evaluation method.

Regarding metrics, we use three correlation coefficients to assess the evaluation consistency between machine-based evaluators and humans: Pearson ($\gamma$)~\citep{mukaka2012pearson}, Spearman ($\rho$)~\citep{zar2005spearman} and Kendall-Tau ($\tau$)~\citep{kendall1938new}.

\begin{table*}[t]
\centering
\resizebox{\textwidth}{!}{%
\begin{tabular}{lcccccccccccc|c}
\toprule
~ & \multicolumn{3}{c}{SummEval} & \multicolumn{3}{c}{TopicalChat} & \multicolumn{3}{c}{SFRES} & \multicolumn{3}{c|}{OpenMEVA (ROC)}& \multirow{2}{*}{Average}\\
\cmidrule(lr){2-4} \cmidrule(lr){5-7} \cmidrule(lr){8-10} \cmidrule(lr){11-13}  ~& $\gamma$ & $\rho$ & $\tau$ & $\gamma$ & $\rho$ & $\tau$ & $\gamma$ & $\rho$ & $\tau$ & $\gamma$ & $\rho$ & $\tau$ \\
\midrule
\multicolumn{14}{c}{\large{Fine-tuning LLM}}\\
\midrule

InstructScore   & 0.3496    & 0.2703    & 0.203     & 0.2691    & 0.2774    & 0.2423    & 0.2039    & 0.1502    & 0.133     & 0.2936    & 0.2772    & 0.1658    & 0.2363 \\

Auto-J & 0.1345    & 0.1457    & 0.1149    & 0.4681    & 0.459     & 0.3714    & 0.1315     & 0.1053    & 0.0869    & 0.3896    & 0.3704    & 0.3065    & 0.257  \\

TIGERScore & 0.458     & 0.3694    & 0.2937    & 0.3785    & 0.4401    & 0.3458    & 0.1898    & 0.1246    & 0.1075    & 0.451     & 0.4413    & 0.3356    & 0.3279 \\

UniEval & 0.5457    & 0.4914    & 0.3707    & 0.5133    & 0.5448    & 0.4134    & 0.3247    & 0.2791   & 0.2081   & 0.4501    & 0.4408    & 0.3119    & 0.4078 \\


\midrule
\multicolumn{14}{c}{\large{Prompting Open-source LLM}}
\\

\midrule

GPTScore-src (FLAN-T5)             & 0.4043    & 0.3584    & 0.2696    & 0.2313    & 0.2437    & 0.1792    & 0.2819     & 0.2082     & 0.1618    & 0.2283    & 0.2265    & 0.1534    & 0.2456 \\

Zero-shot (LLaMA3-8B)   & 0.4104           &  0.3857           &  0.2809      &  {0.5197} &  0.5242      &  0.4018          &  {0.2138}&  0.196         &  0.152                      &  0.4141 &  0.3676  &  0.2808  &   0.3456\\

Zero-shot (Orca2-13B) & \underline{0.5447}    & 0.4916    & 0.3999    & 0.5542    & 0.5512    & 0.4476     & 0.3068     & 0.23      & 0.1842    & 0.4809    & 0.4695    & 0.358     & 0.4182   \\

\textbf{Ours:} \\

\textsc{AC-Coarse} (LLaMA3-8B) &  {0.5307} &  {0.4972}&  {0.3958}&  0.4873    &  {0.5246} &  {0.4259}  &  0.1853         &  0.1594         &  0.1451     &  { 0.4394} &  {0.4403}  &        {0.3477 }  &  {0.3816}   \\

\textsc{AC-Fine} (LLaMA3-8B)   &   {0.5334}    &   {0.502}    &   {0.401}   &   {0.5321 } &   {0.5379}& {0.4045}&   {0.2265}  &   {0.2245 }  &   {0.169 } &   {0.4506}&   {0.4436}  &   \underline{0.3625} &   {0.399 }  \\

\cdashlinelr{2-14}

\textsc{AC-Coarse} (Orca2-13B) & 0.5386    &\underline{0.5227} &  \underline{0.4156} & \textbf{0.611}   & \underline{0.6173} & \textbf{0.4845} &\textbf{0.3612} &\textbf{0.2981}   &\textbf{0.2393}  & \underline{0.4908}  & \underline{0.4962}& {0.3622}& \underline{0.4531 }\\

\textsc{AC-Fine} (Orca2-13B)   &   \textbf{0.6301} & \textbf{0.5486} & \textbf{0.4299 } & \underline{0.6023} & \textbf{0.6214} & \underline{0.4713} & \underline{0.324} & \underline{0.2834} & \underline{0.2289 } & \textbf{0.5259}   & \textbf{0.5363}   &\textbf{0.4109}    &   \textbf{0.4677}  \\

\midrule
\multicolumn{14}{c}{\large{Prompting Close-source LLM}}
\\

\midrule

G-eval (GPT-3.5)                    & 0.4687          & 0.4504             & 0.3745          & 0.5427            & 0.5597           & 0.4501          &    0.2464          &  0.1956          &  0.1591          &  0.362           
                   &  0.3408    &  0.1982   &  0.3624 \\
Zero-shot (GPT-3.5)  & 0.453           & 0.385               & 0.292           & 0.5503            & 0.5436           & 0.4231          &   0.2823          &  0.2274          &  0.1828           &   0.4229        
                   &  {0.397}&  {0.3}&   0.3716  \\
Zero-shot (GPT-4)     & 0.5943             & 0.5038             & 0.4055               & 0.6659             & 0.656          & 0.4937              &  0.3301            &   {0.2823}  &  0.2284      &    \underline{0.5627} & 0.4928         & 0.3777            &   0.4661 \\

\textbf{Ours:} \\

\textsc{AC-Coarse} (GPT-3.5) &\underline{0.6569} &  {0.5368} & {0.4178}& {0.6425}& {0.6171}& {0.4855}& {0.3585} &  \underline{0.2846}  &   \underline{0.2374 }   & {0.4185} &  0.3766  &  0.2981    & {0.4442} \\

\textsc{AC-Fine} (GPT-3.5) &  {0.653} & \textbf{0.6016}   &\textbf{0.4745}   &   {0.6718}    &   {0.6703 }  &   {0.5156 }   &   \underline{0.3616}   & {0.2833 } & {0.2342}&    {0.4693}  &   {0.4527}&   {0.3442} &   {0.4777} \\

\cdashlinelr{2-14}

\textsc{AC-Coarse} (GPT-4)  &  {0.6561} &   {0.5371} &   {0.4277} &  \underline{0.7264} &  \textbf{0.7815} &  \underline{0.6133} &  {0.343} & \textbf{0.2878}      & \textbf{0.2395} &  0.5366          &  \underline{0.5226}  &  \underline{0.4039}   &   \underline{0.5063}   \\

\textsc{AC-Fine} (GPT-4)    & \textbf{0.6926}   & \underline{0.5723 }   &\underline{0.462}      & \textbf{0.7789}&   \underline{0.7753} & \textbf{0.6212} &\textbf{0.363 }    &  0.2809               &  {0.236}& \textbf{0.5877}  & \textbf{0.5581} & \textbf{0.4249} & \textbf{0.5294} \\







\bottomrule
\end{tabular}
}
\caption{Correlation between LLM-based unified evaluators and human judgments on overall quality per instance across four NLG tasks. All train-free evaluators are built upon Orca2-13B. We compare Pearson ($\gamma$), Spearman ($\rho$) and Kendall-Tau ($\tau$) correlation, respectively. The best performance per indicator is highlighted in bold, and the second-highest results are underlined. We implemented and tested all the methods with p-value $<$ 0.05. }

\label{tab:main-result-orca2} 
\end{table*}

\paragraph{Meta-evaluation} 

We establish \method using four widely adopted backbone models: two open-source LLMs (Orca2-13B and LLaMA3-8B) and two closed-source LLMs (GPT-3.5 and GPT-4)\footnote{We used GPT-3.5-turbo-1106 and gpt-4-turbo version for the experiments.} across four diverse NLG tasks. We test two variants of \method (AC) in this study: (1) \textbf{AC-Coarse} performs a coarse-grained, explainable evaluation by prompting the LLM to infer task-related information and directly produce an overall score along with an explanation for each test case. This process considers all inferred criteria at once during scoring alignment. (2) \textbf{AC-Fine} provides a fine-grained, explainable evaluation. Similar to AC-Coarse, it begins with task inference, but during scoring alignment, it assesses the input test case against each criterion individually, offering detailed explanations for each score. The overall quality score is then generated by combining the evaluations across all criteria. Appendix~\ref{parameter} provides the details of implementation.

\section{Results and Analysis}
\label{sec:results}
\begin{table*}[ht]
\centering
\resizebox{0.95\textwidth}{!}{
\begin{tabular}{lcccccccccccc|c}
\toprule
 ~ & \multicolumn{3}{c}{SummEval} & \multicolumn{3}{c}{TopicalChat} & \multicolumn{3}{c}{SFRES} & \multicolumn{3}{c|}{OpenMEVA (ROC)}& \multirow{2}{*}{Average}\\
\cmidrule(lr){2-4} \cmidrule(lr){5-7} \cmidrule(lr){8-10} \cmidrule(lr){11-13}  ~& $\gamma$ & $\rho$ & $\tau$ & $\gamma$ & $\rho$ & $\tau$ & $\gamma$ & $\rho$ & $\tau$ & $\gamma$ & $\rho$ & $\tau$ \\
\midrule

\textbf{Ours (AC-Fine)} & \textbf{0.6301} & \textbf{0.5486} & \textbf{0.4299} & \underline{0.6023} & \textbf{0.6214}& \underline{0.4713} &\underline{0.324}&\underline{0.2834 }   &\underline{0.2289}  &\textbf{0.5259} &\textbf{0.5363}  & \textbf{0.4109}&\textbf{0.4677} \\

\midrule
w/o Task Description &0.5825&  0.4826 & 0.3552 & 0.4949&0.5057&0.4211           &0.2683&0.2017&     0.168          &0.3846&0.3802&0.2918&0.3781\\

w/o Criteria Definition &0.5726 & 0.522  & 0.4062&0.5533&0.5368&0.4451          &0.293&0.2715&      0.1907          &0.4176&0.4237&0.326&0.4132\\

w/o McS-E & 0.5386 & \underline{0.5227} & \underline{0.4156} & \textbf{0.611} & \underline{0.6173}& \textbf{0.4845} &\textbf{0.3612} & \textbf{0.2981} & \textbf{0.2393 }&0.4908&\underline{0.4962}  & 0.3622&\underline{0.4531}\\

w/o OS-E &\underline{0.6106} & 0.5129 & 0.3908&0.5639&0.5615&0.4464             &0.3165&0.2405&0.1899           &\underline{0.509}&0.4931&\underline{0.3632}&0.4332\\

\bottomrule
\end{tabular}
}
\caption{Ablation study of key modules in \method.}

\label{tab:multi-stage_Kendall-Tau}
\vspace{-2mm}
\end{table*}

\subsection{How well does \method perform?}

Table \ref{tab:main-result-orca2} displays the correlation between unified evaluators and human judgments. Overall, \method noticeably outperforms the corresponding prompting-based baselines and state-of-the-art fine-tuning-based evaluators, where our variants built on Orca2-13B and GPT-4 achieve the highest correlation in the methods using open- and close-source LLM, respectively. Comparing two variants of \method per LLM, we find that the fine-level variant consistently achieves higher alignment with human scores, outperforming the coarse one by $\sim$2\% in average correlation. These results show that our approach can effectively enhance LLMs' potential to capture human-centric assessment nuances in diverse scenarios and make more human-aligned judgments. Moreover, prompting the LLM to assess each criterion individually and then aggregate the scores benefits \method's decision-making. 

We also validate the generalizability of \method's self-inferred evaluation prompts on unseen data within a similar NLG scenario. We use the prompts generated by \method on SummEVAL examples to evaluate responses from the Newsroom~\citep{grusky2018newsroom} benchmark, similarly focusing on news article summarization but using diverse strategies. As shown in Appendix~\ref{Generalizes}, our approach outperforms state-of-the-art baselines, demonstrating strong generalization.

Further examining our approach's stability across base LLMs, we observe that \method consistently achieves a noticeable improvement, with an average gain of $\sim$6.8\% correlation over the zero-shot baseline for each base model. This indicates its effectiveness, regardless of the chosen base LLM. Although \method generally obtains greater enhancements when employing a stronger base LLM, it is noteworthy that \method built on Orca2-13B performs comparably to its GPT-4 counterpart on SFRES and OpenMEVA (ROC). Considering the computational cost and evaluator performance, we primarily focus on the ORCA2-based AC-Fine for further analysis.

\subsection{Ablation Study}
\begin{figure}[H] 
\vspace{-3mm}
    \centering  
    \includegraphics[width=0.45\textwidth]{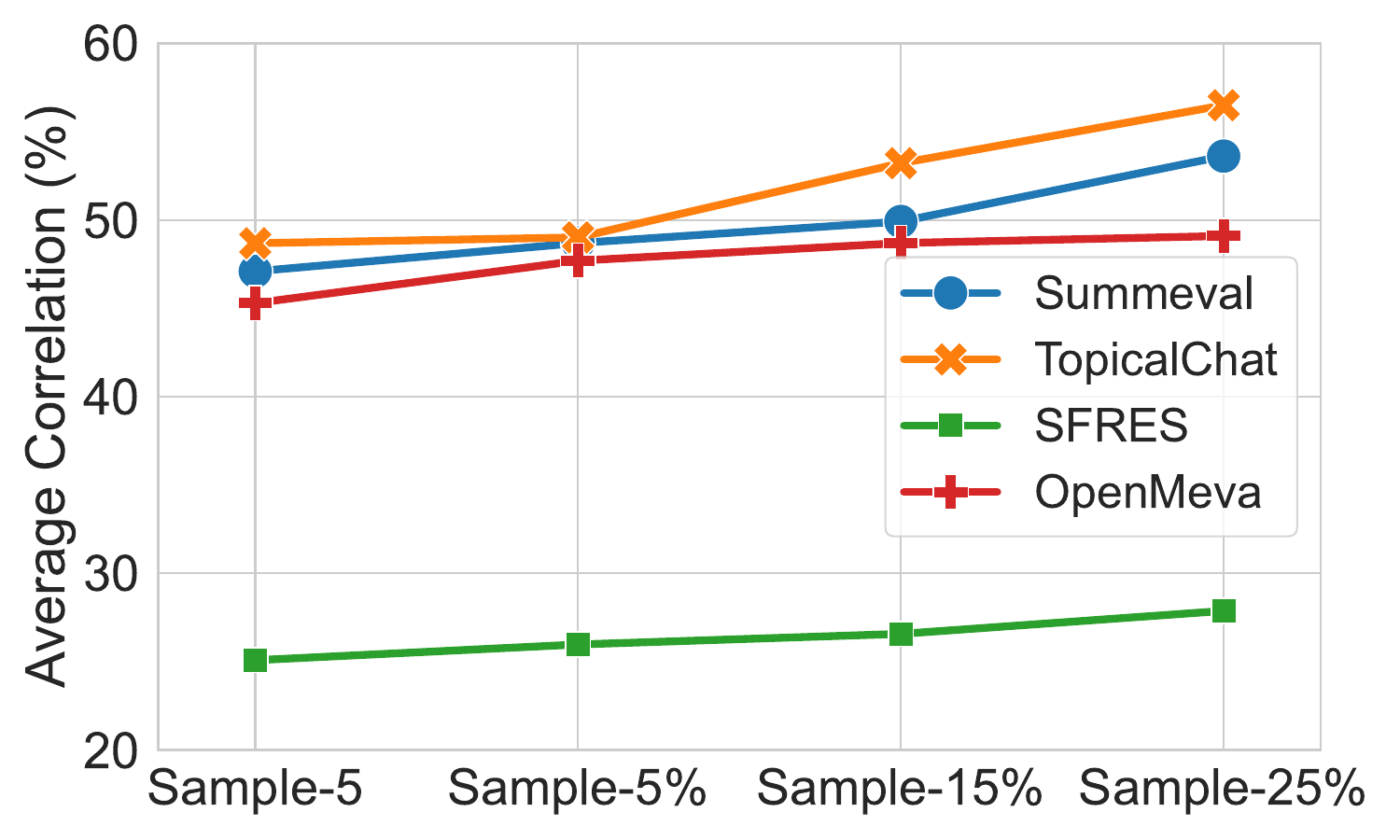} 
    \vspace{-2mm}
    \caption{Average correlation between Orca2-based \method and human judgments with varying label sizes. Results for each correlation coefficient are provided in Appendix~\ref{sec:appendix-trainingsize}} 
    %
    \label{fig:trainingsize} 
\end{figure}

\paragraph{Dependence on Human-scored Data. } To examine the impact of labeled data size on \method's performance, we varied the size of the feed examples from 5-shot to 5\%, 15\%, and the full 25\% of each benchmark. Figure~\ref{fig:trainingsize} shows the results. While \method improves as the labeled data size increases, it can achieve a decent correlation with human evaluators using as few as five human-rated examples. Among four tasks, \method is more sensitive to labeled data size in TopicalChat and SummEval than in OpenMeva and SFRES. The former two benchmarks involve longer contexts and diverse topics, while the latter focus on specific topics with shorter contexts, making the first two tasks more complex. Our observations suggest that \method requires more labeled data for evaluating complex NLG tasks compared to simpler ones.

\begin{figure}[H]
    \vspace{-2mm}
  \centering
  \includegraphics[width=\linewidth]{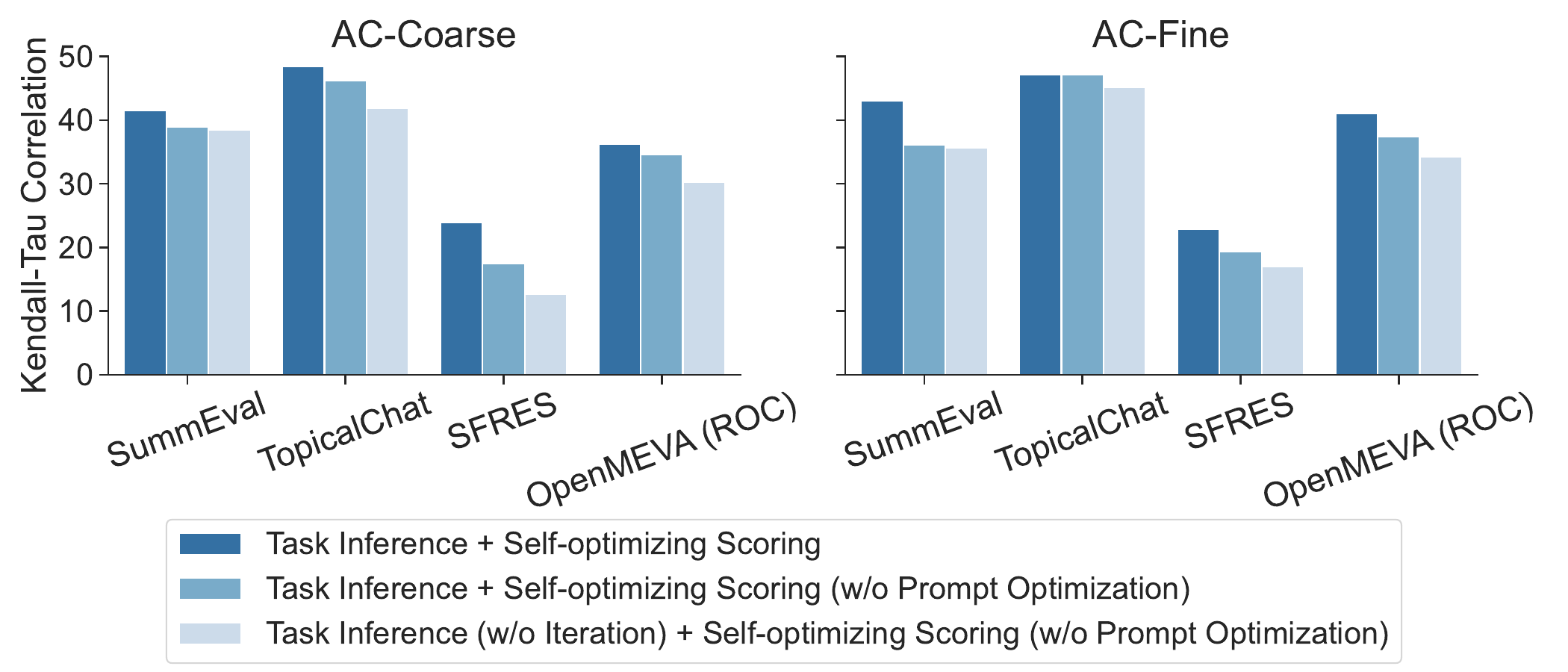}  
  \vspace{-3mm}
  \caption{Impact of prompt optimization on scoring and mini-batch iterations on task inference (Kendall-Tau \%). See Appendix~\ref{sec:appendix-optimization} for Pearson and Spearman results.}
  \label{fig:effectiveness_of_optimization}
\end{figure}

\paragraph{Impact of Optimization.} We compare \method's performance by removing its dynamic prompt optimization for scoring and, furthermore, eliminating mini-batch iterations during task inference. As shown in Figure \ref{fig:effectiveness_of_optimization}, there is a drop in \method's performance when removing scoring prompt optimization, with a further decline when only using a single mini-batch of labeled data for task inference, suggesting that both strategies contribute to \method for making optimal decisions. Interestingly, the \method shows greater sensitivity to scoring optimization in the fine-level evaluation of SummEval and the coarse-level evaluation of SFRES, indicating that this component plays a more significant role in these specific evaluation scenarios. In contrast, the influence of mini-batch iterations for task inference is minimal in SummEval, suggesting that \method can effectively infer the target evaluation task in this setting with limited training data.

\begin{table*}[htbp]
    \centering
    \resizebox{\textwidth}{!}{
\begin{tabular}{>{\raggedright\arraybackslash}p{0.71\linewidth}|>{\raggedright\arraybackslash}p{0.39\linewidth}}
\toprule
\footnotesize
\textbf{Coherence}: The degree to which the summary flows logically and cohesively, with clear connections between the main points.

\textcolor{blue}{\textbf{Conciseness}}: The ability of the summary to convey all necessary information in a succinct and efficient manner. 

\textcolor{blue}{\textbf{Coverage}}: The extent to which the summary captures the main events and details from the source text without omitting crucial information. 

\textcolor{blue}{\textbf{Accuracy}}: The faithfulness of the summary to accurately reflect the main points and details of the source text. 

\textbf{Fluency}: The readability and naturalness of the language used in the summary, with smooth transitions between ideas and paragraphs. 

\textbf{Relevance}: The relevance of the summary to the main topic and the inclusion of only pertinent information from the source text. 

\textcolor{blue}{ \textbf{Clarity}}: The clarity and comprehensibility of the summary, with clear and precise language used to convey the main points. 

\textcolor{blue}{\textbf{Engagement}}: The ability of the summary to captivate and engage the reader, drawing them into the main events and details effectively. 
   
   &
\footnotesize   
\textbf{Coherence}: the summary should be well-structured and well-organized. The summary should not just be a heap of related information, but should build from sentence to sentence to a coherent body of information about a topic.

\textbf{Consistency}: the factual alignment between the summary and the summarized source. A factually consistent summary contains only statements that are entailed by the source document.

\textbf{Fluency}: the summary should have no formatting problems, capitalization errors or obviously ungrammatical sentences (e.g., fragments, missing components) that make the text difficult to read.

\textbf{Relevance}: The summary should include only important information from the source document. \\ \bottomrule
\addlinespace[-17pt]
\multicolumn{1}{c}{\textbf{\scriptsize (a) AC-Fine }}&
\multicolumn{1}{c}{\textbf{\scriptsize (b) Human}}
\\
\end{tabular}
}
    \caption{
    An illustrative example of the generated evaluation criteria on SummEval, either generated by an \method (a) or predefined by humans (b). The highlighted text in blue are additional criteria generated by the machine compared to the human-defined ones.}
    \label{tab:critriavis}
\vspace{-2mm}
\end{table*}

\paragraph{Module Contribution.} Table~\ref{tab:multi-stage_Kendall-Tau} shows the individual contribution of each module in \method. Note that the variant w/o criteria inference uses the original predefined criteria from each benchmark for further computation. In the variant w/o OS-E, we calculated the overall quality score per test case by averaging the multiple criteria-specific scores generated by McS-E. We find that removing task inference modules leads to a more substantial performance drop compared to removing scoring modules, especially when the LLM is not asked to infer the task description (resulting in a $\sim$9\% decrease on average). Our findings suggest that guiding LLMs to understand what to evaluate contributes more to \method's effectiveness than other modules. Additionally, the larger performance drop in the variant w/o OS-E, compared to the one w/o McS-E, indicates that the LLM-generated overall quality score contributes more meaningfully than simply averaging the criteria-specific scores.

\subsection{Qualitative Analysis of \method}

\paragraph{LLM-inferred Criteria  Analysis. } Moving forward from quantitative analysis, we examine the LLM-inferred criteria in depth. Table~\ref{tab:critriavis} shows an illustrative comparison between the criteria generated by \method and those pre-defined by humans in SummEval. We find that our approach incorporates more nuanced criteria (i.e., ``clarity'', ``conciseness'',  ``coverage'', and ``engagement'') beyond the four pre-defined aspects. Moreover, each criterion is paired with a clear definition to specify its distinct characteristics. For example, the human-defined ``coherence'' starts with a high-level description like ``well-structured and well-organized'', while the LLM's definition tends to be more concrete, e.g., ``the summary flows logically''.

\begin{table}[t]
\centering
\resizebox{\linewidth}{!}{
\begin{tabular}{ccccc}
\toprule
 \textbf{Dimension} & 
 \textbf{Clarity}  & \textbf{Relevance} & \textbf{Score Consistency}& \textbf{Accuracy} \\ 
 Rate$\rightarrow$ & Yes (\%) & Yes (\%) & Yes (\%) & Yes (\%) 
 \\
 
 \midrule
Coherence & 99.11  & 92  & 95.78 & 85.33 \\ 

 Conciseness & 98.67 & 91.78  & 96.89 & 88.89\\ 
 
Coverage & 98.82  & 91.33  & 97.56 & 96.89 \\ 

Accuracy & 98.22  & 92.22  & 95.56 & 98\\ 

 Fluency & 99.56  & 98.89  & 96 & 96.67 \\ 
 
Relevance & 98.89  & 99.11 & 98.44 & 95.56 \\ 

 Clarity & 98  & 94.22  & 93.56 & 95.78 \\ 
 
 Engagement & 99.33  & 94.67  & 93.33 & 91.11\\ 
 
 Overall Quality & 98.44  & 98.44  & 97.33 & 98 \\ 
 \midrule
Average & 98.78 & 94.74 & 96.05& 94.03\\ 
 \bottomrule
\end{tabular}
}
\vspace{-2mm}
\caption{Human evaluation of criterion-specific explanations on SummEval samples.}
\label{tab:human Evaluation for criterion-specific Explanation}
\end{table}

\begin{table}[t]
\centering
\resizebox{\linewidth}{!}{
\begin{tabular}{ccm{2.8cm}c}
\toprule
 \textbf{Dimension} &\textbf{Aspect-to-Overall Alignment} & \textbf{Differentiability} & \textbf{Usefulness} \\ 
 Rate$\rightarrow$ & Yes (\%) & Yes (\%) & (1-5)
 \\
 
 \midrule
 Overall& 95.11  & 90 & 4.515 \\ 
 \bottomrule
\end{tabular}
}
\vspace{-2mm}
\caption{Human evaluation of overall explanations on SummEval samples, emphasizing (1) the alignment of the overall explanation with criterion-specific ones, (2) explanations' differentiability across vary-quality cases, and (3) explanations' overall usefulness per case.}

\vspace{-5pt}
\label{tab:human Evaluation for overall Explanation}
\end{table}
\paragraph{Human Evaluation of Explanations. } We also employ three proficient English-speaking annotators to evaluate the quality of the scoring explanations generated by \method on a random sample of 150 test cases from SummEval (see details in Appendix~\ref{huaman eval}).  As shown in Table \ref{tab:human Evaluation for criterion-specific Explanation}, the individual explanations demonstrate comparatively high quality across four dimensions, with average scores of 98.78\% for clarity, 94.74\% for relevance, 96.05\% for score consistency, and 94.03\% for information accuracy. As shown in Table~\ref{tab:human Evaluation for overall Explanation}, the overall explanations generally align with the criteria-specific ones (95.11\%), and 90\% of the overall explanations effectively differentiate case quality. With an average rating of $\sim$4.5 out of 5 on the generated explanations across sampled testing cases, the result shows that explanations generated by \method are of good quality and useful to explain the resulting scores. 

\section{Conclusion}
\label{sec:conclusion}

We proposed \method, a novel LLM-based NLG evaluation protocol that relies solely on lightweight human-scored data. Unlike existing machine-based evaluators that depend on human-predefined task-related information for assessment, \method self-identifies the target evaluation task and nuanced evaluation criteria purely from the data for making judgments. This paradigm shift will enhance the adaptability of \method, enabling it to flexibly capture the varying priority expectations of different end-users across diverse generation scenarios. Our approach reduces the need for intensive manual efforts to design task-specific criteria and extensive prompt engineering. Experiments across four distinct NLG tasks demonstrate LLMs' potential as active critics, achieving higher correlation with human judgments compared to baselines. Fine-level criteria-specific scoring, paired with explanations, prompts the LLM to engage more deeply with the test cases, leading to improved overall quality scoring.

\newpage
\section{Limitation}
\label{sec:limitation}

Our work has several limitations. First, due to resource constraints, we primarily focused on four existing NLG tasks and benchmarks for meta-evaluation in our experiments. It would be valuable to deploy our protocol in a broader testing environment to assess its performance in more diverse settings. Additionally, building \method on a wider range of backbone LLMs could provide deeper insights. Overall, we hope this study will contribute to advancing generic NLG evaluation research and promote system development across diverse NLG scenarios.

\bibliography{custom}

\onecolumn
\newpage
\appendix

\section{An example of evaluation protocol and prompt on SummEval}
\label{Appendix-promptexample}

\subsection{An example of input data}
This section shows an example of data $(x_i, y_i, r_i)$ from SummEval in Table~\ref{tab:appendix-dataexample}.

\begin{table*}[htbp]
    \centering
    \begin{tabular}{>{\raggedright\arraybackslash}p{15cm}}
        \toprule
        \textbf{Source ($x_i$)}\\
        \midrule
        A southern Iowa chiropractor accused of accepting sex as payment for his services and performing exorcisms on patients has surrendered his state license. The Iowa Board of Chiropractic released a report Wednesday detailing charges against Charles Manuel, of Lamoni. Manuel signed an agreement last month admitting his misdeeds and pledging not to apply for reinstatement for at least 10 years. Patient satisfaction: A chiropractor in Iowa has surrendered his license to practice and admitted to swapping services for sex and performing exorcisms on some patients. Nonetheless, he's received outstanding evaluations for patient satisfaction on HealthGrades.com The agreement requires that Manuel prove any circumstances surrounding the revocation of his license no longer exist before he can resume practicing chiropractic in the state. Those circumstances included bartering sex for services with some patients. Manuel also recommended that patients stop taking medication he prescribed to them. A woman who answered a call to Manuel's home from the Des Moines Register declined to comment on the case. A woman at his former practice said he hadn't worked there for some time. A Lamoni address listed on the Yelp page for Manuel's practice appears to be a home on a residential street. While Maneul has received just three patient survey responses on HealthGrades.com, those responses were quite positive. The disgraced chiropractor received a perfect five out of five stars in patient satisfaction. Strange Practice: Charles Manuel, who admitted wrongdoing to the Iowa Board of Chiropractic, listed his practice's addresses on this residential street in the small agricultural town of Lamoni in southern Iowa.
        \\
        \midrule
        \textbf{Response ($y_i$)}\\
        \midrule
        The iowa board of chiropractic released a report wednesday detailing charges against charles manual, of lamoni. Manuel signed an agreement last month admitting his misdeeds and pledging not to apply for reinstatement for at least 10 years . The agreement requires that manuel prove any circumstances surrounding the revocation of his license no longer exist before he can resume practicing chiropractic in the state.
        \\
        \midrule
        \textbf{Human Judgment Score ($r_i$)}\\
        \midrule
        88.33333333333332\\
        \midrule  

    \end{tabular}
    \caption{An example of data $(x_i, y_i, r_i)$ from SummEval}
    \label{tab:appendix-dataexample}
\end{table*}

\clearpage

\subsection{ An Example of Output}
This section shows an example of output with our AC-Fine method in table~\ref{tab:appendix-outputexample}.
\nopagebreak
\begin{table}[htbp]
  \centering
  \resizebox{\linewidth}{!}{
  \begin{tabular}{>{\raggedright\arraybackslash}p{19cm}}
\toprule

\textbf{Multi-Criteria Scoring with Explanation}\\
\midrule
Coherence: \{\\
\hspace*{0.6cm}    explanation: The response is coherent and logically flows from one point to the next, with clear connections between the main points.\\
\hspace*{0.6cm}    score: 90\\
\}\\
Conciseness: \{\\
\hspace*{0.6cm}    explanation: The response is concise and efficiently conveys the main events and details from the source text.\\
\hspace*{0.6cm}    score: 90 \\
\}\\
Coverage: \{\\
\hspace*{0.6cm}    explanation: The response covers the main events and details from the source text without omitting crucial information.\\
\hspace*{0.6cm}    score: 85  \\
\}\\
Accuracy: \{\\
\hspace*{0.6cm}    explanation: The response accurately reflects the main points and details of the source text.\\
\hspace*{0.6cm}    score: 90  \\
\}\\
Fluency:\{\\
\hspace*{0.6cm}    explanation: The response is fluent, with smooth transitions between ideas and paragraphs, and natural language used throughout.\\
\hspace*{0.6cm}    score: 95\\
    \}\\
Relevance: \{\\
\hspace*{0.6cm}    explanation: The response is highly relevant to the main topic and includes pertinent information from the source text.\\
\hspace*{0.6cm}    score: 95\\
\}\\
Clarity: \{\\
\hspace*{0.6cm}    explanation: The response is clear and comprehensible, with precise language used to convey the main points effectively.\\
\hspace*{0.6cm}    score: 90\\
\}\\
Engagement: \{\\
\hspace*{0.6cm}    explanation: The response effectively draws the reader into the main events and details, making it highly engaging.\\
\hspace*{0.6cm}    score: 95\\
\}\\
\midrule
\textbf{Overall Scoring with Explanation}\\
\midrule
\textbf{Explanation:} The response provides a concise and coherent summary of the source text, capturing the main events and details without omitting crucial information. The summary is accurate, relevant, and clear, with smooth transitions between ideas and paragraphs, making it highly engaging and readable. The coverage of the main events and details is comprehensive, and the language used is clear and precise. Overall, the response effectively conveys the main points of the source text in a compelling and engaging manner.
\\
\textbf{Overall Score:} 90.0\\
\bottomrule
  \end{tabular}
}
\caption{An example of output with our AC-Fine method} 
\label{tab:appendix-outputexample}
\end{table}
\vspace*{-3cm}

\clearpage

\subsection{ Instruction component of the evaluation protocol}

This section presents the instruction $I$ in evaluation protocol $\Phi$, which is also the output of the task inference module, in table~\ref{tab:appendix-instruction}

\begin{table*}[htbp]
  \centering
  \begin{tabular}{>{\raggedright\arraybackslash}p{15cm}}
\toprule
\textbf{Task Description $T$:}\\
\midrule
The evaluation task is text summarization, which refers to evaluating abstractive summaries generated from source texts. The relationship between `Source' and `Response' is that the responses consistently provide a summary of the main events or details described in the source text and accurately reflect the main points of the source text in a summarized form.\\
\midrule
\textbf{Critiria $C$:}\\
\midrule

``Coherence": ``The degree to which the summary flows logically and cohesively, with clear connections between the main points."

``Conciseness": ``The ability of the summary to convey all necessary information in a succinct and efficient manner." 

``Coverage": ``The extent to which the summary captures the main events and details from the source text without omitting crucial information." 

``Accuracy": ``The faithfulness of the summary to accurately reflect the main points and details of the source text." 

``Fluency": ``The readability and naturalness of the language used in the summary, with smooth transitions between ideas and paragraphs." 

``Relevance": ``The relevance of the summary to the main topic and the inclusion of only pertinent information from the source text." 

``Clarity": ``The clarity and comprehensibility of the summary, with clear and precise language used to convey the main points." 

``Engagement": ``The ability of the summary to captivate and engage the reader, drawing them into the main events and details effectively." \\

\bottomrule
  \end{tabular}
\caption{An example of instruction $I$ in evaluation protocol $\Phi$} \label{tab:appendix-instruction}
\end{table*}

\clearpage

\subsection{In-context exemplar of the evaluation protocol}
This section presents the in-context exemplar $D_{demo}$ in evaluation protocol $\Phi$ in table~\ref{tab:appendix-demo}

\begin{table*}[htbp]
  \centering
  \begin{tabular}{>{\raggedright\arraybackslash}p{15cm}}
\toprule
\textbf{AC-Fine Output Example: }\\
\midrule
\textbf{Source:} ``Paul Merson has restarted his row with Andros Townsend... Any bad feeling between the pair seemed to have passed but Merson was unable to resist having another dig at Townsend after Tottenham drew at Turf Moor."

\textbf{Response:} ``Paul merson has restarted his row with andros townsend .. in the 83rd minutefor tottenham as they drew 0-0 against burnley."

\textbf{Multiple Evaluation Criteria:}

Coherence: The degree to which the summaryflows logically and cohesively, with clearconnections between the main point.

Conciseness: The ability of the summary to convey all necessaryinformation in a succinctand efficient manner.

...

\textbf{Score Of Each Criterion In JSON:} 

Coherence: \{ \\
\hspace*{1cm} Explanation: The response is somewhat coherent, but it jumps between different events and \hspace*{1cm} details without clear connections between them. \\
\hspace*{1cm} Score: 60 \\
 \hspace*{1cm}   \}\\
...\\

\textbf{Explanation: }The response provides a concise summary ... to provide a more compelling and logically flowing summary. \\

\textbf{Score of overall:} 75\\

\midrule
\textbf{$\mathcal{D}_{train}$ Example: }\\
\midrule
\textbf{Source:} Chelsea have made an offer for FC... The initial five-year deal is the biggest in the club 's history , with the Blues now considering a two-week pre-season tour of Japan this summer.\\

\textbf{Response:} Chelsea have made an offer for fc ... in muto is not connected to the 200million sponsorship deal they signed with japanese company yokohama rubber in February.

\textbf{"Score of Overall":} 91.66666666666666\\
\bottomrule
  \end{tabular}
\caption{An example of in-context exemplar $D_{demo}$ } \label{tab:appendix-demo}
\end{table*}

\clearpage

\subsection{Prompt Template}
This section presents prompt templates in multiple stages: (1) Task Description (Table~\ref{tab:appendix-TaskDescription}), (2) Criteria Definition (Table~\ref{tab:appendix-CriteriaDefinition}), (3) Multi-Criteria Scoring with Explanation (Table~\ref{tab:appendix-Multi-CriteriaScoring}), and (4) Overall Scoring with Explanation (Table~\ref{tab:appendix-OverallQualityScoring}).

\begin{table*}[htbp]
  \centering
  \begin{tabular}{>{\raggedright\arraybackslash}p{15cm}}
\toprule
Given several examples from an NLG evaluation dataset where each entry consists of a `Source' text and its corresponding `Response', along with a score that evaluates the response quality.

Please write observations about trends that hold for most or all of the samples.

I will also provide you with some previous observations I have already made.  Please add your observations or if you feel the observations are comprehensive say `COMPLETE'.

Some areas you may consider in your observations: content and structure, scenario, task, evaluation objective, evaluation criteria, etc.

It will be useful to make an educated guess as to the nature of the task this dataset will enable. Don't be afraid to be creative.\\

\$\{\textit{examples}\}

\$\{\textit{prior observations}\}
\\
\midrule

Given a series of observations I have made and some description about this NLG evaluation dataset.

\hspace*{0.6cm}    1. Identify the type of evaluation task. Possible tasks include: machine translation, text summarization, data-to-text generation, dialogue generation, image description, text simplification, story generation, paraphrase generation, textual entailment, reasoning, etc.
    
\hspace*{0.6cm}    2. What this evaluation task refers to evaluating.
    
\hspace*{0.6cm}    3. Output the relationship between `Source' and `Response' in this task in 1-3 sentences.
    
\hspace*{0.6cm}    4. Given a summary in fill [ ]: The evaluation task is [ ], which refers to evaluating [ ] generated from [ ]. The relationship between `Source' and `Response' is [ ].\\

\$\{\textit{observations}\}

\$\{\textit{prior task description}\}
\\
\bottomrule
  \end{tabular}
\caption{Prompt template on Task Description} 
\label{tab:appendix-TaskDescription}
\end{table*}

\begin{table*}[htp]
  \centering
  \begin{tabular}{>{\raggedright\arraybackslash}p{15cm}}
\toprule
Given a task description about this NLG evaluation dataset and a series of observations I have made.

Your task is to list ten aspects that can be considered when measuring the overall quality of \$\{\textit{task type}\}. 

\$\{\textit{task description}\}

\$\{\textit{observations}\}

Output in JSON format: aspect as key, description as value.\\
\midrule

From the provided sets of criteria for evaluating \$\{\textit{task type}\}, identify the key aspects that are essential for this task.  Select between 4 to 10 criteria that best align with the goals of your evaluation task and prioritize them based on their importance to the overall quality of the \$\{\textit{task type}\}. 

\$\{\textit{sets of criteria}\}

Output in JSON format: aspect as key, description as value. \\
\bottomrule
  \end{tabular}
\caption{Prompt template on Criteria Definition} 
\label{tab:appendix-CriteriaDefinition}
\end{table*}

\begin{table*}[htp]
  \centering
  \begin{tabular}{>{\raggedright\arraybackslash}p{15cm}}
\toprule
\$\{\textit{Task Description}\}

Your task is to evaluate the response on multiple evaluation criteria with respect to the source on a continuous scale from 0 to 100, and explain your process for scoring each criterion. Rate the response on multiple evaluation criteria and give a brief explanation in a JSON format by filling in the placeholders in [ ].
\\
\\
\$\{\textit{In-context exemplar}\}
\\
\\
\$\{\textit{Source}\}

\$\{\textit{Response}\}

\$\{\textit{Multiple Evaluation Criteria}\}
\\
\\
Output format:

Score Of Each Criterion In JSON:
\\
\\
\{ \\
Coherence: \{ \\
\hspace*{1cm}    Explanation: ``[your explanation]'', \\
\hspace*{1cm}    Score: ``[score from 0 to 100: 0 - No logic, 100 - Perfectly coherent]'' \}, \\
Conciseness: \{ \\
\hspace*{1cm}    Explanation: ``[your explanation]'', \\
\hspace*{1cm}    Score: ``[score from 0 to 100: 0- Overly verbose, 100- Highly efficient]'' \}, \\
...

\} 

\\
\bottomrule
  \end{tabular}
\caption{Prompt template on Multi-Criteria Scoring with Explanation} \label{tab:appendix-Multi-CriteriaScoring}
\end{table*}

\begin{table*}[htp]
  \centering
  \begin{tabular}{>{\raggedright\arraybackslash}p{15cm}}
\toprule
\$\{\textit{Task Description}\}

Your task is to rate the overall quality of the response, based on the source and the scores for different criteria of the response on a continuous scale from 0 to 100, where 0 means `completely irrelevant and unclear' and 100 means `perfectly relevant, clear, and engaging.' IMPORTANT!! Only output the score as an `int' and nothing else.

``Also explain your process to get this score to response. Also please perform error Analysis of given response. What should we change to have a better result?"
\\
\\
\$\{\textit{In-context exemplar}\}
\\
\\
\$\{\textit{Source}\}

\$\{\textit{Response}\}

\$\{\textit{Score Of Different Criteria}\}
\\
\\
Output format:
\\
\\
Explanation: 
\\
Score Of Overall:
\\

\bottomrule
  \end{tabular}
\caption{Prompt template on Overall Scoring with Explanation} \label{tab:appendix-OverallQualityScoring}
\end{table*}

\clearpage

\section{Details of Parameter Setting and Implementation}
\label{parameter}

We randomly sample 25\% of the data for \method tuning and use the remaining 75\% for meta-evaluation across each NLG task. During task inference, we set the number of mini-batches to 25, with a batch size of 5. The LLM is instructed to generate one task description and a set of evaluation criteria per mini-batch. To enhance tuning efficiency, we allow the LLM to decide when to stop early, capping the number of task descriptions and criteria sets at 5. For the scoring stage, we run 11 epoches of prompt optimization. The number of in-context exemplars used per epoch is 3 for SummEval and TopicalChat, and 8 for SFRES and OpenMeVA (ROC), with the difference due to varying input text lengths across tasks. All parameter settings are based on empirical testing of sequential values to determine optimal configurations.

Our experiments were carried out using two NVIDIA V100 GPU cards. For prompt optimization in the scoring stage, we utilized the ``BootstrapFewShotWithRandomSearch" method in DSPy~\citep{khattab2023dspy} as the optimizer, which leverages random search to generate examples.

\section{Generalization to Unseen Datasets}
\label{Generalizes}

Ideally, we expect the \method-generated evaluation prompts can be directly used for NLG system assessment in a similar future NLG scenario. To assess the generalizability of these prompts, we use the prompts generated by \method based on SummEval examples to assess unseen cases in Newsroom~\citep{grusky2018newsroom}. This dataset comprises 60 news articles and their corresponding summaries generated by 7 summarization systems. Each summary is paired with an overall quality score provided by human annotators. Table~\ref{tab:generalizestonewsroom} displays the results. Our \method noticeably outperforms baselines by $\sim$10\% correlation on average, indicating \method's generalizability.

\begin{table}[ht]
\centering
\resizebox{0.5\linewidth}{!}{
\begin{tabular}{lccc|c}
\midrule
Method & $\gamma$ & $\rho$ & $\tau$  & AVE \\
\midrule
TIGERScore & 0.3731 & 0.41   & 0.3075    & 0.3635\\
UniEval    & 0.4485 & 0.4505 & 0.325   &  0.408 \\
\midrule
G-eval (gpt3.5)     & 0.3853 & 0.4053 & 0.3012  & 0.3639 \\
GPT-3.5 (zero-shot)  & 0.504  & 0.561  & 0.430 &0.4983 \\
\textbf{\textsc{AC-Fine} (GPT3.5)} &\textbf{0.6382}& \textbf{0.6444} & \textbf{0.4949} &\textbf{0.5925} \\
\midrule
GPT-4 (zero-shot) & 0.6583 & 0.6649 & 0.4957 &0.6063\\
\textbf{\textsc{AC-Fine} (GPT4)}  &\textbf{0.7466}&\textbf{0.7111}& \textbf{0.5474}&\textbf{0.6684} \\
\midrule
\end{tabular}
}
\caption{Generalization results of \method on Unseen Datasets. }
\label{tab:generalizestonewsroom}
\end{table}

\section{Additional Results of \method's Dependence on Human-scored Data}
\label{sec:appendix-trainingsize}

\begin{figure}[htbp]
\centering
\includegraphics[width=\textwidth]{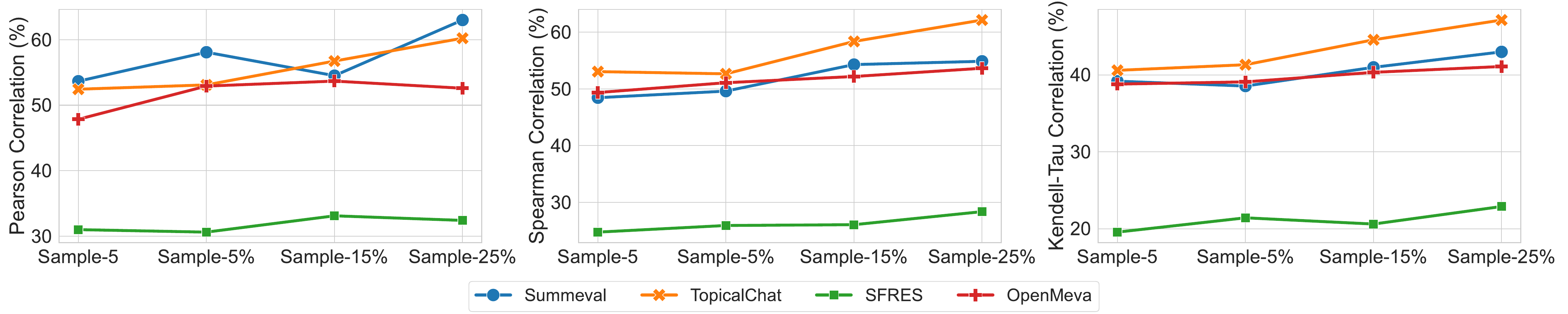}
        \caption{Results of \method's dependence on human-scored data by Pearson, Spearman, and Kendell-Tau, respectively.}
        \label{fig:Pearson-Spearman-Kendell-Tau}
\end{figure}

\section{Impact of Optimization by Pearson and Spearman Correlation}
\label{sec:appendix-optimization}

\begin{figure}[H]
    \centering
    \includegraphics[width=0.8\linewidth]{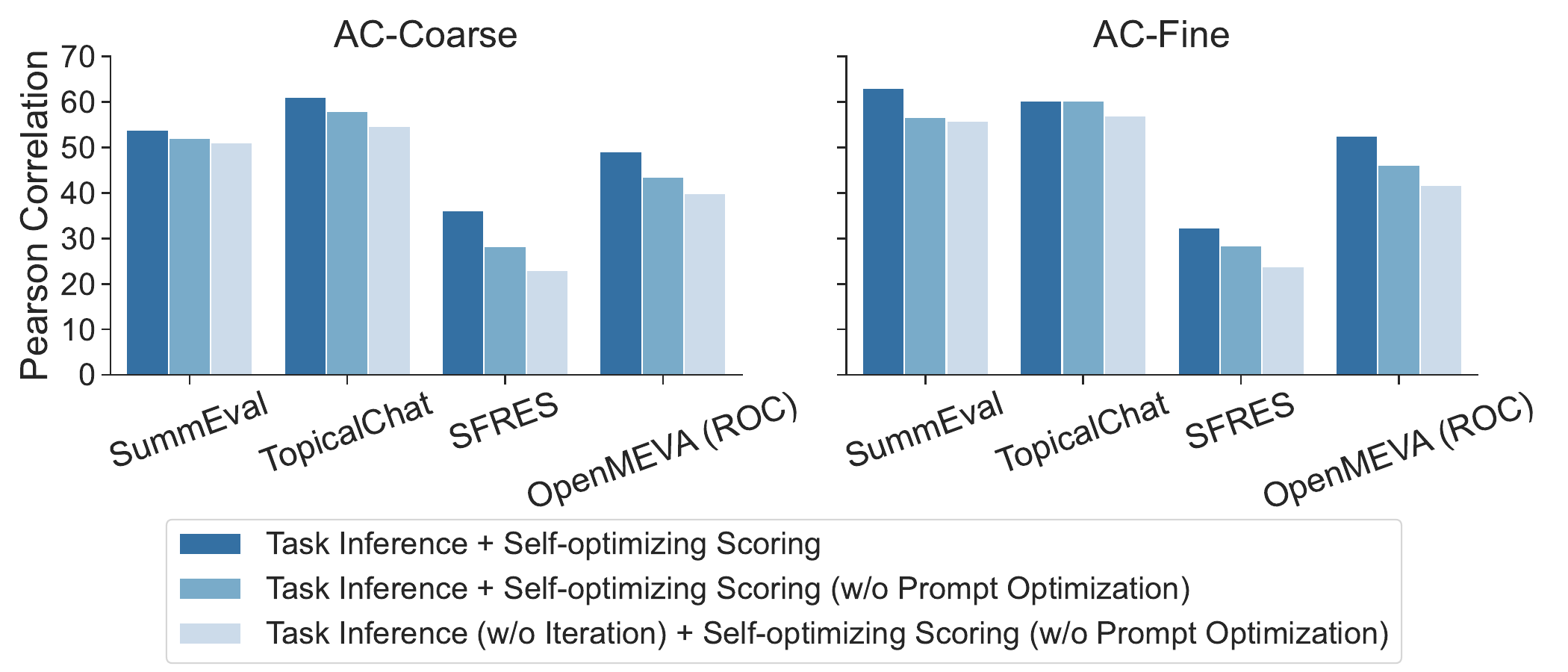}
    \caption{Effectiveness of Optimization. We report the Pearson ($\gamma$) correlation coefficient for our two optimal experimental variants: AC-Coarse and AC-Fine.  }
    \label{fig:effectiveness_of_Optimization_Pearson}
\end{figure}
\vspace{+10mm}
\begin{figure}[H]
    \centering
    \includegraphics[width=0.8\linewidth]{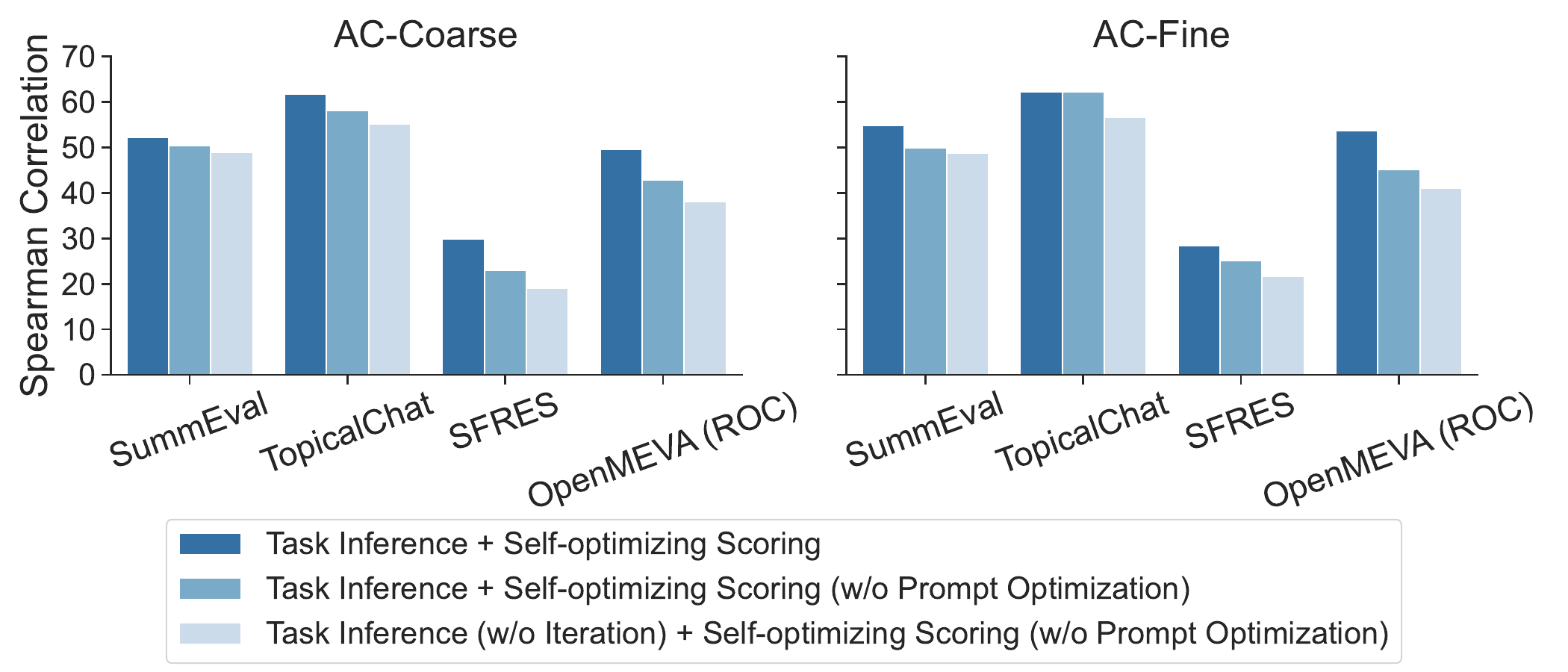}
    \caption{Effectiveness of Optimization. We report the Spearman ($\rho$) correlation coefficient for our two optimal experimental variants: AC-Coarse and AC-Fine.  }
    \label{fig:ffectiveness_of_Optimization_spearman}
\end{figure}

\vspace{+10mm}

\section{Helpfulness of Explanations to \method's Judgments}
\label{Quantitative Analysis of Explanation}

\begin{figure}[H]
  \centering
  \includegraphics[width=0.69\linewidth]{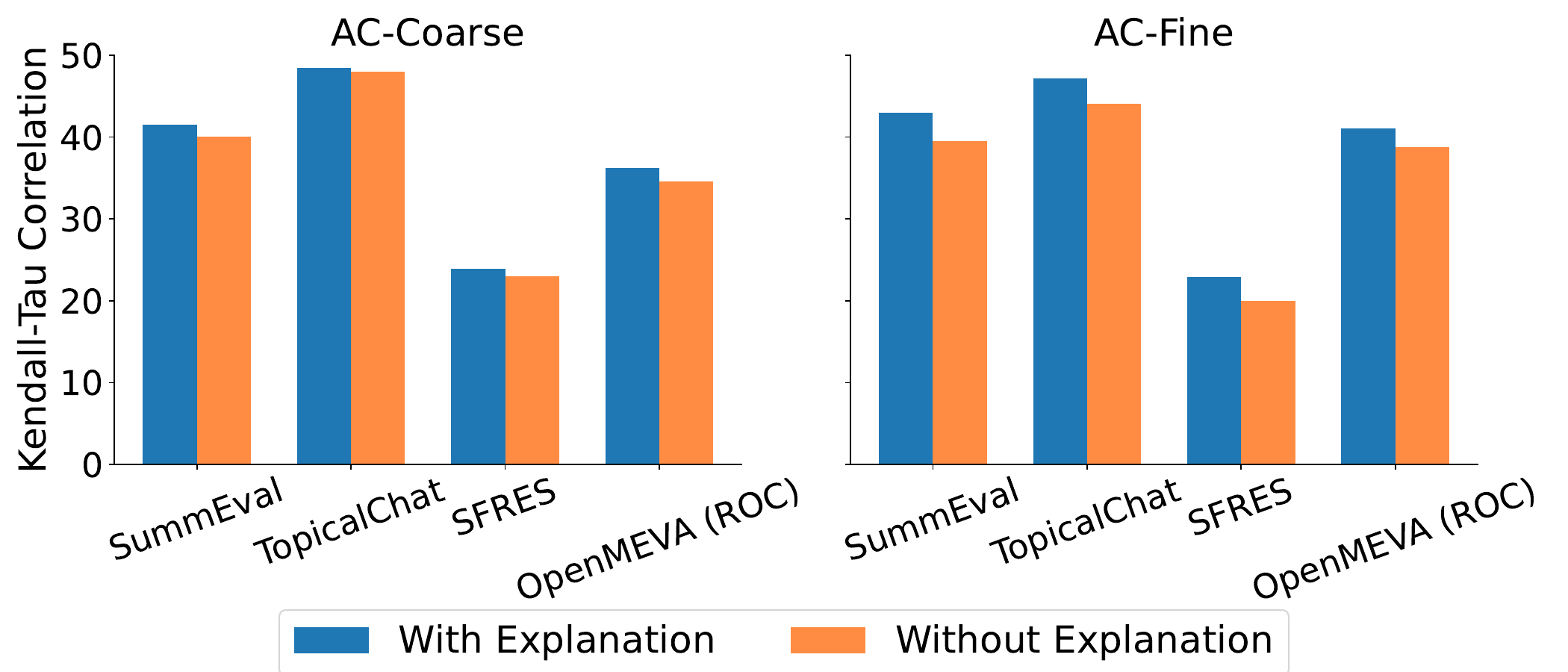}  
  \caption{Comparison of the performance of \method w/ vs. w/o explanations across four NLG tasks. We measure \method's performance by Kendall-Tau correlation (\%). Both coarse-level (AC-Coarse) and fine-level (AC-Fine) variants are investigated.}
    \label{fig:effectiveness_of_explanation1_compiled}
\end{figure}

\begin{figure}[H]
    \centering
    \includegraphics[width=0.69\linewidth]{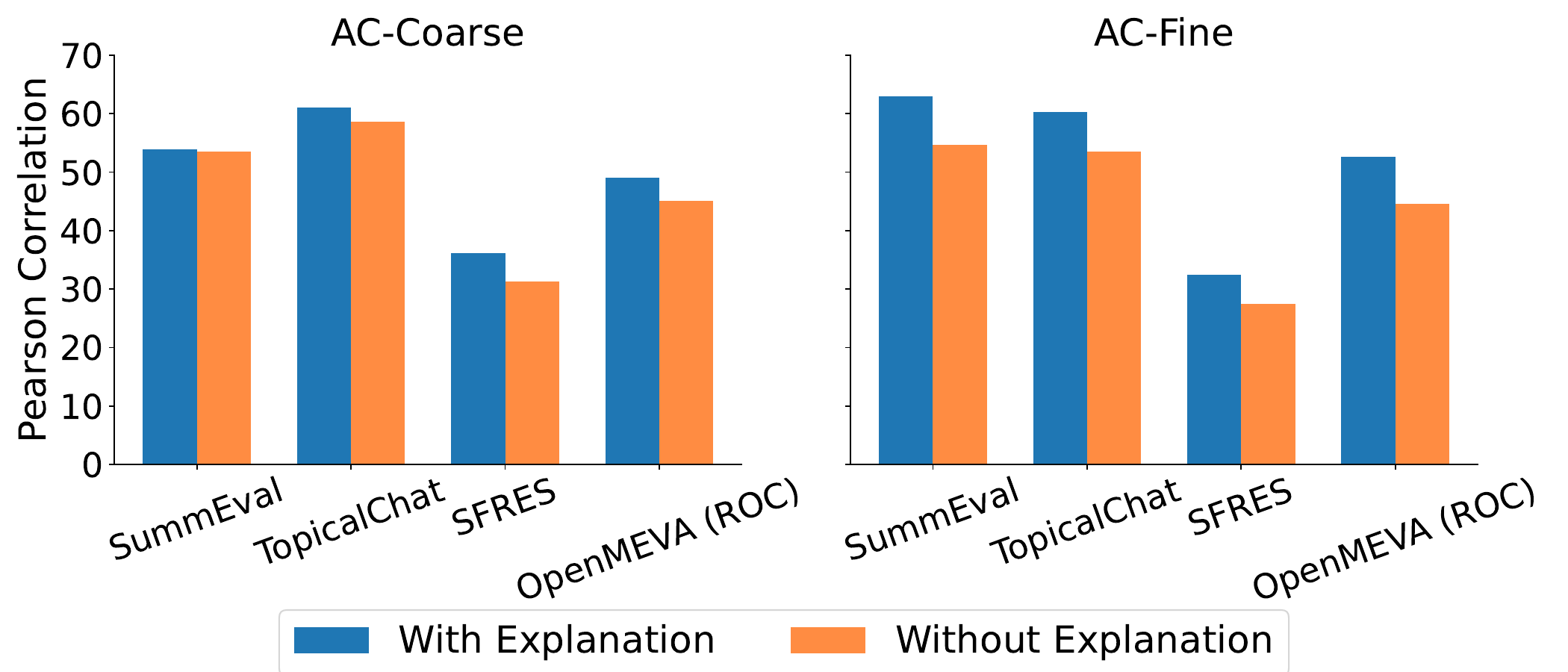}
    \caption{Effectiveness of Explanation in Pearson ($\gamma$).  }
    \label{fig:effectiveness_of_explanation_pearson}
\end{figure}

\begin{figure}[H]
    \centering
    \includegraphics[width=0.69\linewidth]{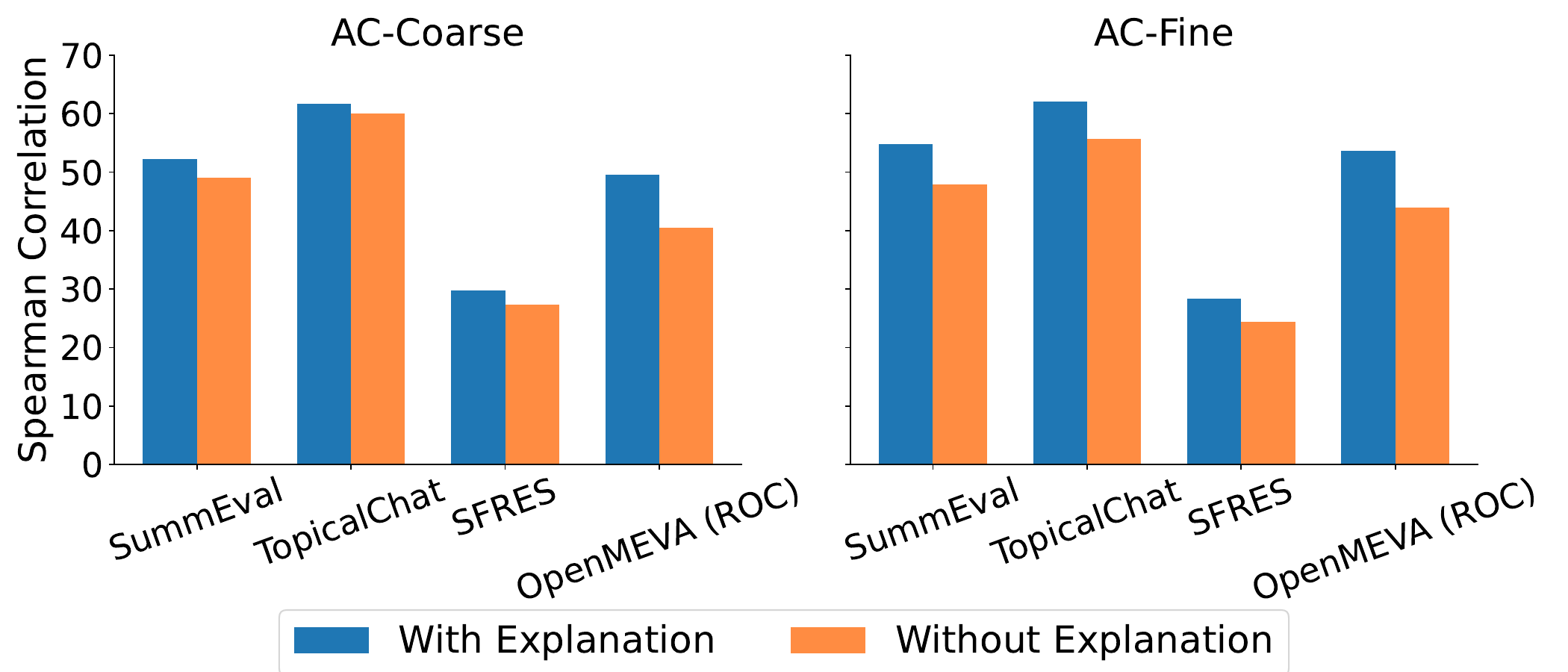}
    \caption{Effectiveness of Explanation in Spearman ($\rho$). }
    \label{fig:effectiveness_of_explanation_spearman}
\end{figure}

To assess the impact of explanations generated by \method, we compared our protocol's performance with versus without explanations, at both coarse and fine levels of evaluations. Figure \ref{fig:effectiveness_of_explanation1_compiled} shows the results based on the Kendall-Tau correlation. We also provide the results of Pearson and Spearman correlation in Figure \ref{fig:effectiveness_of_explanation_pearson} and Figure \ref{fig:effectiveness_of_explanation_spearman} respectively.


As shown in Figures \ref{fig:effectiveness_of_explanation1_compiled}, \method with explanations consistently demonstrates a higher correlation with human judgments than the version without explanations. Notably, the difference in correlation is greater for the fine-level \method compared to the coarse-level variant. These findings suggest that generating explanations for scoring helps the base LLM engage more effectively in the evaluation process, resulting in stronger alignment with human judgments. In particular, fine-level explanations for each model-inferred criterion are especially effective in boosting the model's engagement and improving evaluation accuracy.

\section{Details of Human Evaluation Design on \method's Explanations}\label{huaman eval}

Our assessment consisted of four parts, with details provided below. First, for each individual explanation per case, each annotator rated the quality based on: (1) clarity of the statement, (2) relevance to the target criterion, (3) alignment with the corresponding score, and (4) accuracy within the context of the test case (e.g., correctness in matching the source text). Further emphasizing the overall scoring explanation per case, we asked annotators to assess its alignment with the criteria-specific explanations, and its differentiability across cases of varying quality, respectively. Finally, we asked annotators to provide an overall rating on a scale of 1-5 based on the usefulness of all generated explanations per case. To validate the reliability of human annotations, following prior work ~\citep{fabbri2021summeval}, we calculated intercoder reliability by Krippendorff’s alpha~\citep{krippendorff2011computing}. The 0.6534 Kappa coefficient indicates substantial agreement among annotators.

\clearpage

\begin{figure}[ht]
\centering
\includepdf[pages=1, scale=0.8, pagecommand={}]{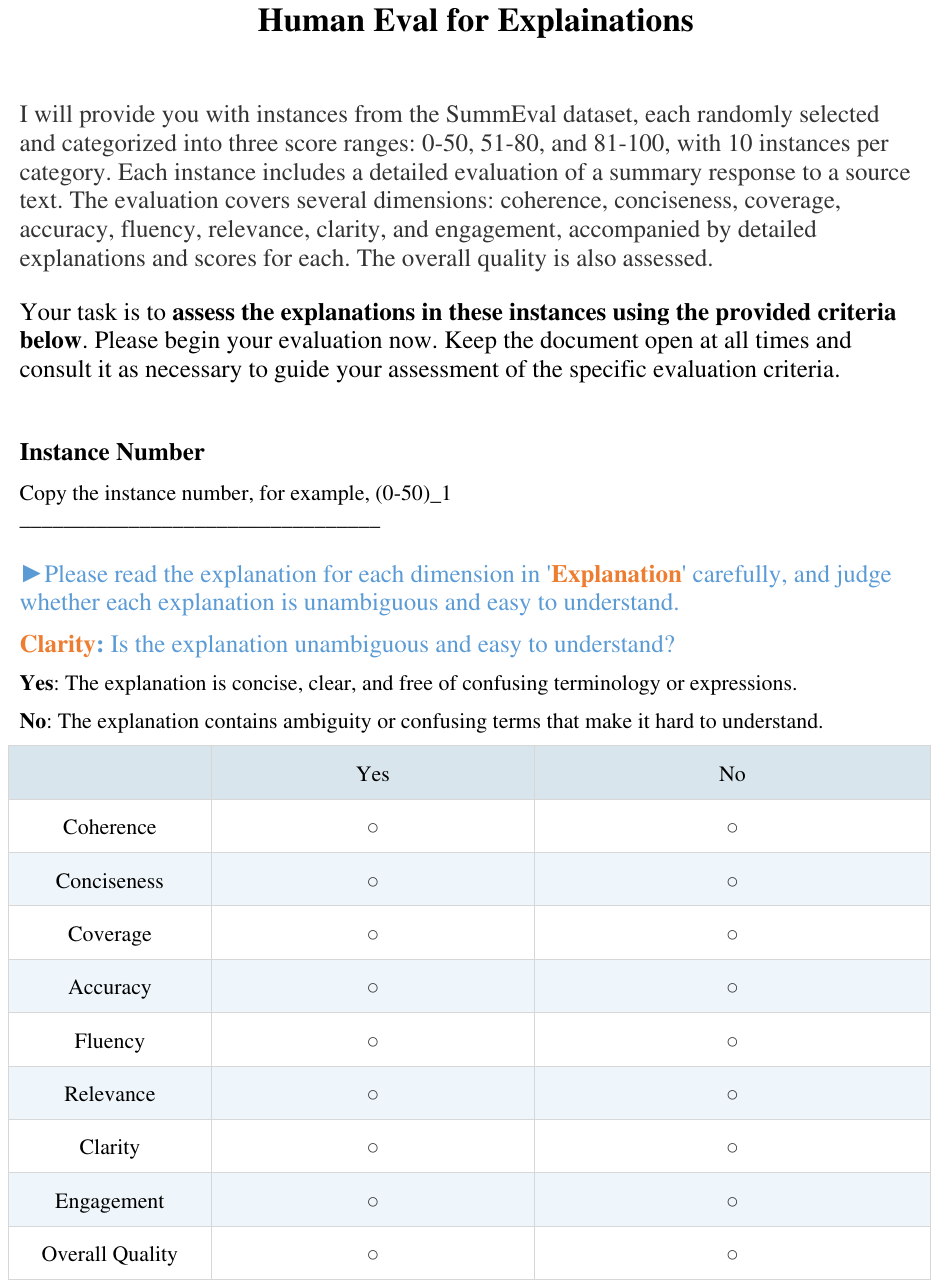}
\end{figure}

\clearpage

\begin{figure}[ht]
\centering
\includepdf[pages=2, scale=0.8, pagecommand={}]{figs/Guidance_Human_Eval_for_Explainations.pdf}
\end{figure}

\clearpage

\begin{figure}[ht]
\centering
\includepdf[pages=3, scale=0.8, pagecommand={}]{figs/Guidance_Human_Eval_for_Explainations.pdf}
\end{figure}

\clearpage

\begin{figure}[ht]
\centering
\includepdf[pages=4, scale=0.8, pagecommand={}]{figs/Guidance_Human_Eval_for_Explainations.pdf}
\end{figure}



\end{document}